\documentclass[11pt]{article}

\usepackage[]{acl}

\usepackage{times, graphicx, footmisc, layouts}
\usepackage{latexsym,multirow}

\usepackage[utf8]{inputenc}

\usepackage{microtype}

\newcommand{\data}{\textsc{XWikiRef}}
\newcommand{\task}{\textsc{XWikiGen}}

\title{XWikiGen: Cross-lingual Summarization for Encyclopedic Text Generation in Low Resource Languages}

\author{Dhaval Taunk, Shivprasad Sagare, Anupam Patil, Shivansh Subramanian,\\ {\bf Manish Gupta and Vasudeva Varma} \\ Information Retrieval and Extraction Lab, IIIT Hyderabad, India \\ \{dhaval.taunk,shivprasad.sagare,shivansh.s\}@research.iiit.ac.in,anupampatil44@gmail.com\\ \{manish.gupta,vv\}@iiit.ac.in}

\begin{document}
\maketitle
\begin{abstract}
Lack of encyclopedic text contributors, especially on Wikipedia, makes automated text generation for \emph{low resource (LR) languages} a critical problem.
Existing work on Wikipedia text generation has focused on \emph{English only} where English reference articles are summarized to generate English Wikipedia pages.
But, for low-resource languages, the scarcity of reference articles makes monolingual summarization ineffective in solving this problem. 
Hence, in this work, we propose \task{}, which is the task of cross-lingual multi-document summarization of text from multiple reference articles, written in various languages, to generate Wikipedia-style text. 
Accordingly, we contribute a benchmark dataset, \data{}, spanning $\sim$69K Wikipedia articles covering five domains and eight languages. We harness this dataset to train a two-stage system where the input is a set of citations and a section title and the output is a section-specific LR summary. The proposed system is based on a novel idea of neural unsupervised extractive summarization to coarsely identify salient information followed by a neural abstractive model to generate the section-specific text.
Extensive experiments show that multi-domain training is better than the multi-lingual setup on average. 
We make our code and dataset publicly available\footnote{\url{https://doi.org/10.5281/zenodo.7604438}\label{datafootnote}}.
\end{abstract}

\section{Introduction}
\begin{figure}
    \centering
    \includegraphics[width=\columnwidth]{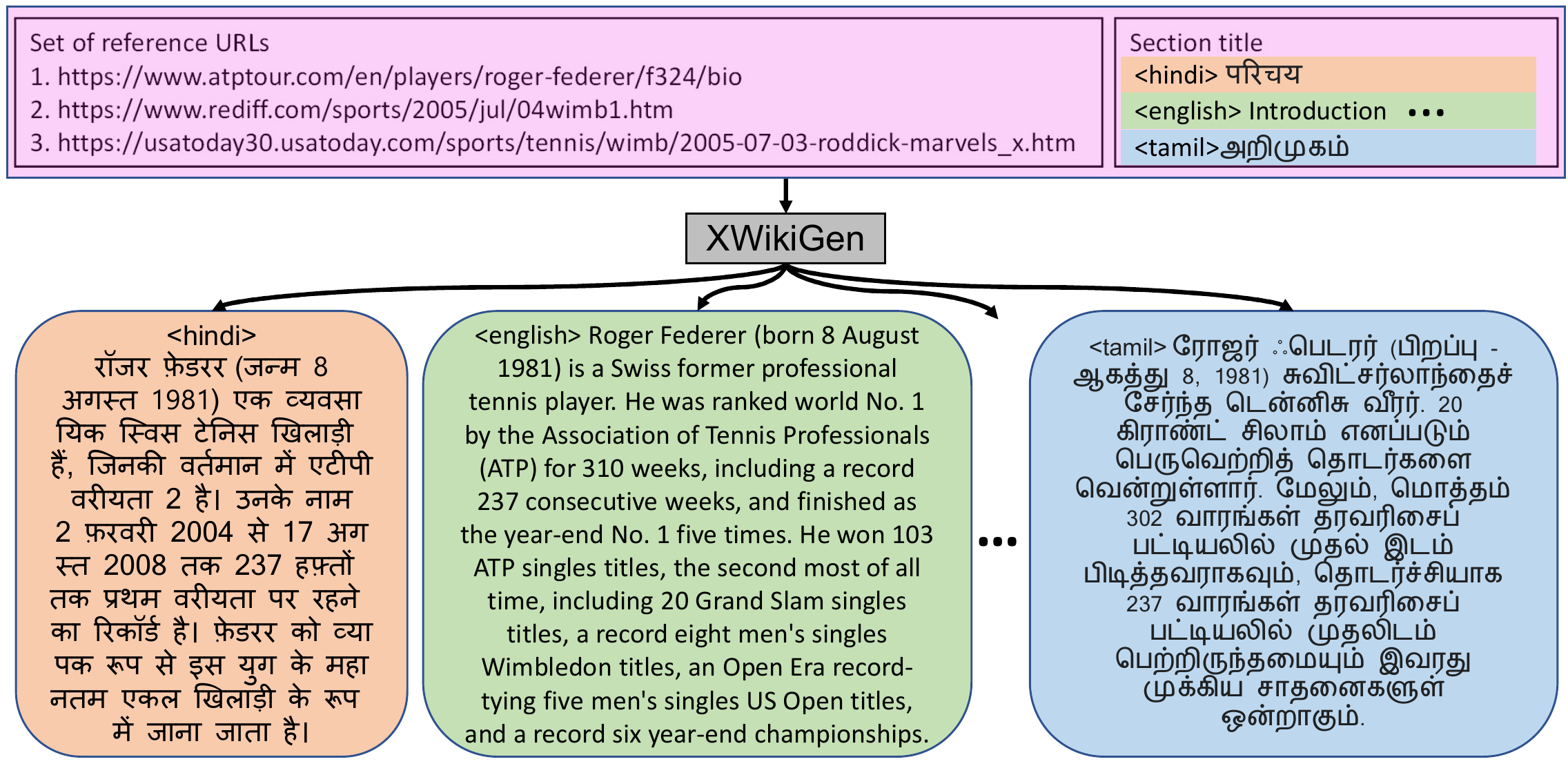}
    \caption{\task{} examples: Generating Hindi, English, and Tamil text for the Introduction section from cited references.}
    \label{fig:examples}
\end{figure}



\begin{figure}[!t]
    \centering
    \includegraphics[width=\columnwidth]{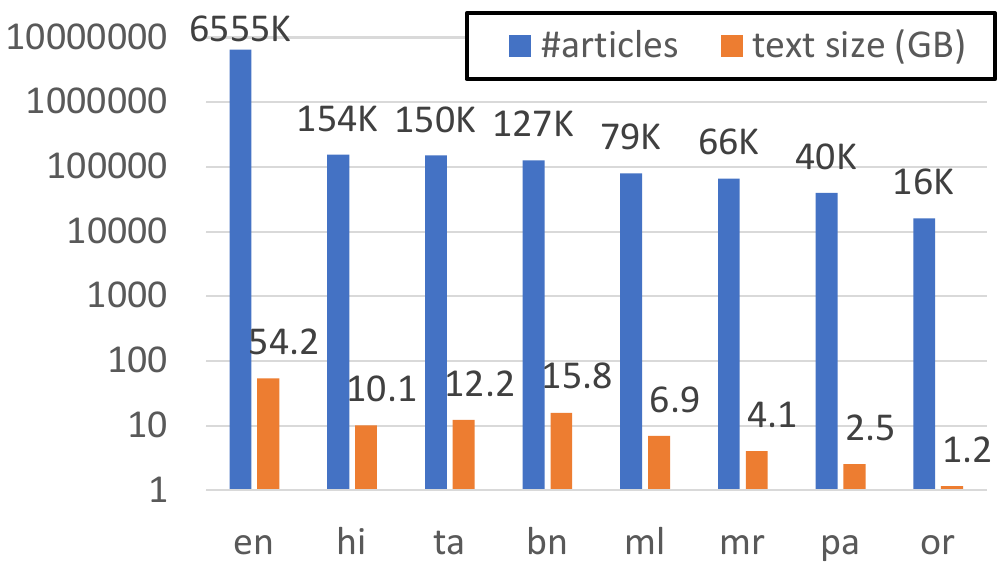}
    \caption{Number of Wikipedia articles and text size in GBs across eight languages, using 20220926 Wikipedia dump. Note that the Y axis is in log scale.}
    \label{fig:wikiStats}
\end{figure}

\begin{figure}[!b]
    \centering
    \includegraphics[width=\columnwidth]{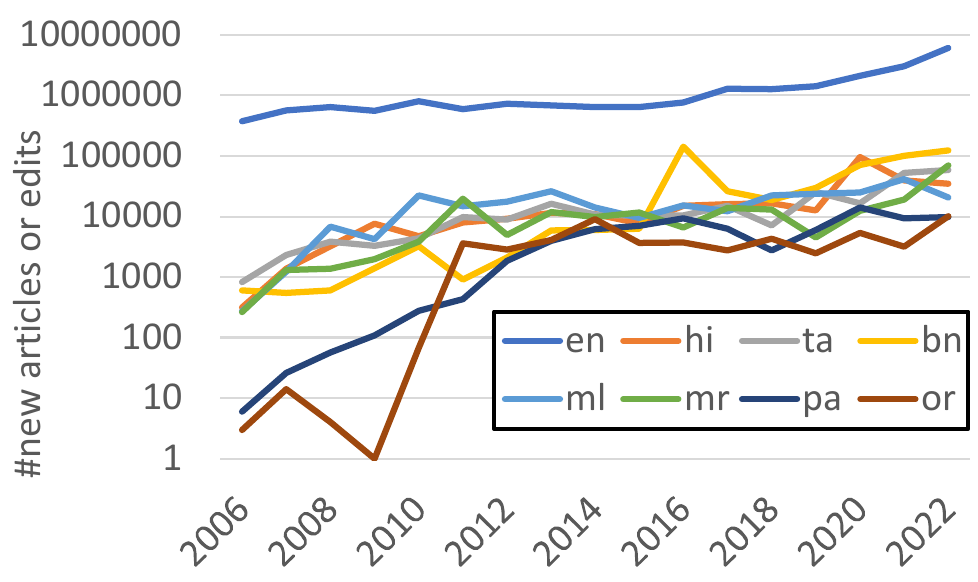}
    \caption{Number of new articles or edits on Wikipedia across eight languages from 2006 to 2022. This is obtained using a publication date from the 20220926 Wikipedia dump. Note that the Y axis is in the log scale.}
    \label{fig:wikiRateStats}
\end{figure}

Although Wikipedia has been the primary choice of encyclopedic reference for millions of users, unfortunately, Wikipedia is extremely sparse for low-resource (LR) languages. English Wikipedia exhibits abundance with $\sim$6.56M articles expressed in 54.2 GB of text, low resource Wikipedia is poor with only $\sim$90K articles expressed using 7.5 GB of text on average across seven low resource languages as shown in Fig.~\ref{fig:wikiStats}. Further, as illustrated in Fig.~\ref{fig:wikiRateStats}, manual efforts towards enriching LR Wikipedia over the years have also not been as encouraging as in the case of English. These observations indicate that automated text generation for low-resource Wikipedia is critical.


A possible na\"ive approach for automated generation for low-resource Wikipedia is to translate text from equivalent English Wikipedia articles. Unfortunately, several low-resource entities of interest tend to be local in nature, leading to a lack of equivalent English Wikipedia pages for $\sim$42.1\% entities on average across seven low-resource languages. In particular, following are percentages of Wikipedia entities with no equivalent English Wikipedia page: Hindi (50.60\%), Tamil (46.70\%), Bengali (31.50\%), Malayalam (36.30\%), Marathi (42.00\%), Punjabi (38.70\%), Oriya (39.40\%). Thus, we need to explore other inputs for LR Wikipedia text generation.

Another approach is to leverage generic Web content for LR Wikipedia text generation. A challenge to this approach is that such web content is itself very sparse in low-resource languages, as can be observed in publicly available large dumps like CommonCrawl~\cite{raffel2020exploring}. Further, we analyzed the language of cited references on existing Wikipedia pages for eight languages and five domains of our interest using langdetect library\footnote{\url{https://pypi.org/project/langdetect/}}. From Table~\ref{tab:refLang}, we observe that while English Wikipedia pages have more than 85\% references in English, the numbers are minuscule for most (domain, LR language) pairs. This motivates us to propose a novel problem of cross-lingual Wikipedia text generation using references, which we call \task{} in short. 

\begin{table}
    \centering
    \small
    \begin{tabular}{|l|c|c|c|c|c|c|c|c|}
    \hline
Domain&bn&hi&ml&mr&or&pa&ta&en\\
    \hline
    \hline
books&16.5&14.9&9.9&12.3&0.0&5.2&28.2&94.8\\
\hline
films&21.5&10.4&21.0&6.5&0.0&1.2&10.9&96.8\\
\hline
politicians&21.4&31.2&8.4&25.0&0.0&1.9&8.7&90.0\\
\hline
sportsmen&1.4&1.7&1.2&2.5&0.0&0.2&1.1&87.2\\
\hline
writers&11.0&18.3&4.6&27.2&0.0&6.0&7.7&94.7\\
\hline
    \end{tabular}
    \caption{Percentage of cited references with language same as Wikipedia article language (for 8 languages and 5 domains which are a part of our \data{} dataset).}
    \label{tab:refLang}
\end{table}




As shown in Fig.~\ref{fig:examples}, the input for \task{} is a set of reference URLs, a target section title, and a target output language. The expected output is then the text suitable for that Wikipedia section in the target language. Analogous to generic summarization versus query-based summarization, \task{} involves section-wise text generation rather than the generation of the entire Wikipedia page. Unlike existing work on monolingual (English-only) Wikipedia text generation, \task{} is cross-lingual in nature. Lastly, unlike some existing work that generates cross-lingual text using English Wikipedia pages, \task{} focuses on generating cross-lingual text using reference URLs in multiple languages.


Our first contribution is a novel dataset, \data{} towards the \task{} task. The dataset is obtained from Wikipedia pages corresponding to eight languages and five domains. Languages include Bengali (bn), English (en), Hindi (hi), Malayalam (ml), Marathi (mr), Oriya (or), Punjabi (pa) and Tamil (ta). Domains include books, films, politicians, sportsmen, and writers. The dataset spans $\sim$69K Wikipedia articles with $\sim$105K sections. Each section has 5.44 cited references on average. 


\task{} is an extremely challenging task because it involves long text generation, and that too in a cross-lingual manner. Handling long text input is difficult. Hence, we follow a two-stage approach. The first extractive stage identifies important sentences across several reference documents. The second abstractive stage generates the section text. Both stages involve neural models. We experiment with unsupervised methods like salience~\cite{yasunaga2021qa} and hiporank~\cite{dong2020discourse} for the extractive stage, and mT5~\cite{xue2021mt5} and mBART~\cite{liu2020multilingual} for the abstractive stage. We experiment with several training setups like (1) multi-lingual, (2) multi-domain, and (3) multi-lingual-multi-domain. We report results using standard text generation metrics like ROUGE-L, METEOR, and chrF++. We make our code and dataset publicly available.

Overall, we make the following contributions in this paper.
\begin{itemize}
    \item We motivate and propose the \task{} problem where the input is (set of reference URLs, section title, language) and the output is a text paragraph.
    \item We contribute a large dataset, \data{}, with $\sim$105K instances covering eight languages and five domains.
    \item We model \task{} as a multi-document cross-lingual summarization problem and propose a two-stage extractive-abstractive system. Our multi-lingual-multi-domain models using HipoRank (extractive) and mBART (abstractive) lead to the best results. 
\end{itemize}


\setlength{\tabcolsep}{2pt}
\begin{table*}
    \centering
        \small
    \begin{tabular}{|p{1in}|p{1.2in}|c|c|c|c|c|p{1in}|p{1.4in}|}
    \hline
Dataset&\#Summaries&XL?&ML?&\#Langs&MD?&SS?&Input&Output\\
    \hline
        \hline
        WikiSum~\cite{liu2018generating}&$\sim$2.3M articles&No&No&1&Yes&No&Set of citation URLs&Whole Wiki article\\
\hline
WikiAsp~\cite{hayashi2021wikiasp}&$\sim$400K sections&No&No&1&Yes&Yes&Set of citation URLs&One section in same language\\
\hline
GameWikiSum~\cite{antognini2020gamewikisum}&$\sim$26K gameplay Wikipedia sections&No&No&1&Yes&No&Professional video game reviews&Gameplay Wikipedia sections\\
\hline
Wiki Current Events Portal (WCEP)~\cite{ghalandari2020large}&$\sim$10.2K WCEP event summaries&No&No&1&Yes&No&Set of news articles&WCEP Summary\\
\hline
MultiLing'15~\cite{giannakopoulos2015multiling}&$\sim$1.5K paragraphs&No&Yes&38&No&No&Whole Wikipedia article&First few Wikipedia sentences in same language\\
\hline
WikiMulti~\cite{tikhonov2022wikimulti}&$\sim$150K intro paragraph&Yes&Yes&15&No&No&Whole Wikipedia article&Intro paragraph in other language\\
\hline
XWikiRef (Ours)&$\sim$105K sections&Yes&Yes&8&Yes&Yes&Set of citation URLs&One section in another language\\
\hline
    \end{tabular}
    \caption{Statistics of popular Wikipedia Summarization datasets. XL=Cross-lingual. ML=Multi-Lingual. MD=Multi-document. SS=Section-specific.}
    \label{tab:datasetComparison}
\end{table*}

\section{Related Work}
\label{sec:relatedWork}
In this section, we discuss related work on generating both short and long Wikipedia text. We also briefly discuss work on multi-lingual and cross-lingual summarization.
\subsection{Generating Short Wikipedia Text}
Automated generation of Wikipedia text has been a problem of interest for the past 5--6 years. Initial efforts in the fact-to-text (F2T) line of work focused on generating short text, typically the first sentence of Wikipedia pages using structured fact tuples. 

Training F2T models require aligned data with adequate content overlap. Some previous studies like WebNLG~\cite{gardent2017webnlg} collected aligned data by crowdsourcing, while others have performed automatic alignment by heuristics like TF-IDF. Seq-2-seq neural methods~\cite{lebret2016wikibio,mei2016talk} have been popularly used for F2T. These include vanilla LSTMs~\cite{vougiouklis2018neural}, LSTM encoder-decoder model with copy mechanism~\cite{shahidi2020two}, LSTMs with hierarchical attentive encoder~\cite{nema2018generating}, pretrained Transformer based models~\cite{ribeiro2021investigating} like BART~\cite{lewis2020bart} and T5~\cite{raffel2020exploring}.

Most of the previous efforts on F2T focused on English fact-to-text only. Only recently, the Cross-lingual F2T (XF2T) problem was proposed in~\cite{abhishek2022xalign} and~\cite{sagare2022xf2t}. Compared to all of these pieces of work which have focused on short text generation, the focus of the current paper is on generating longer text. Unlike F2T literature, where the input is structured, the input is a set of reference URLs in our case.

\subsection{Generating Long Wikipedia Text}
Besides generating short Wikipedia text, there have also been efforts to generate Wikipedia articles by summarizing long sequences~\cite{liu2018generating,ghalandari2020large,hayashi2021wikiasp,antognini2020gamewikisum,giannakopoulos2015multiling,tikhonov2022wikimulti} as shown in Table~\ref{tab:datasetComparison}. For all of these datasets, the generated text is either the full Wikipedia article or text for a specific section. Most of these studies~\cite{liu2018generating,hayashi2021wikiasp,antognini2020gamewikisum,ghalandari2020large} have been done on English only. Further, these studies use different kinds of input: single document (existing Wikipedia article in the same or another language) or multi-document (set of citation URLs, review pages). 


Most of these works fail to include the section-specific intent during summarization and generate an article on the whole. Hence, to capture the section-specific intent while summarizing, Hayashi et al.~\cite{hayashi2021wikiasp} introduced section-specific summarization, which recognizes the main topics in the input text and then creates a summary for each. Although authors rely on the model to figure out the latent subtopics, the content selection step is challenging. We tackle this challenge by providing section-specific citations as input in our dataset, which avoids the noisy references belonging to other sections, and allows us to study the summarization capabilities of the model better.

Interestingly, none of the existing datasets perform cross-lingual multi-document summarization for Wikipedia text. However, as motivated in the previous section, this setup is critical for Wikipedia text generation for LR languages. Hence, we fill this gap in this paper by proposing the \data{} dataset.


\subsection{Multi-lingual and cross-lingual summarization}
Recently there has been a lot of work on multi-lingual and cross-lingual NLG tasks like machine translation~\cite{chi2021mt6,liu2020multilingual}, question generation~\cite{chi2020cross,mitra2021zero}, news title generation~\cite{liang2020xglue}, blog title generation~\cite{bhatt2023multilingual}, and  summarization~\cite{zhu2019ncls,jhaveri2019clstk} thanks to models like XNLG~\cite{chi2020cross}, mBART~\cite{liu2020multilingual}, mT5~\cite{xue2021mt5}, etc.

Limited work has been done in the past on summarization for low-resource languages. MultiLing'15~\cite{giannakopoulos2015multiling} introduced a novel task for multi-lingual summarization in 30 languages. In the past 2--3 years, a few datasets have been proposed for cross-lingual summarization mainly in the news domain: XLSum~\cite{hasan2021xl}, MLSum~\cite{scialom2020mlsum}, CrossSum~\cite{hasan2021crosssum}, Global Voices~\cite{nguyen2019global}, WikiLingua~\cite{ladhak2020wikilingua}, WikiMulti~\cite{tikhonov2022wikimulti}. XL-Sum~\cite{hasan2021xl} comprises $\sim$1.35 million professionally annotated article-summary pairs from BBC, extracted using a set of carefully designed heuristics. It  covers 44 languages ranging from low to high resource. Hasan et al.~\cite{hasan2021crosssum} extend the multi-lingual XL-Sum dataset by releasing CrossSum, a cross-lingual summarization dataset with $\sim$1.7 million instances. However, both XL-Sum and CrossSum are specific to the news domain only. WikiLingua~\cite{ladhak2020wikilingua} is a multi-lingual dataset with $\sim$770K summaries where the article and summary pairs are extracted in 18 languages from WikiHow. MLSum and GlobalVoices are also cross-lingual summarization datasets based on news articles with around $\sim$1.5M and $\sim$300K summaries covering 5 and 15 languages, respectively. We enrich this line of work by contributing a new cross-lingual multi-document summarization dataset, \data{}, and also proposing a two-stage system for the associated \task{} task. 


\section{\data{}: Data Collection, Pre-processing and Analysis}
\label{sec:dataset}
In this section, we first discuss the procedure for \data{} data collection and pre-processing. Then we present detailed analysis.


\subsection{Data Collection and Pre-processing}
\data{} contains Wikipedia sections related to five distinct domains (books, films, politicians, sportsmen, writers) spanning across eight languages (bn, en, hi, ml, mr, or, pa, ta). We start by using Wikidata API\footnote{\url{https://query.wikidata.org/}} to filter the domains of interest initially and further fetch the entities that have Wikipedia pages in our set of languages. Later, we use Wikipedia language-specific 20220926 XML dumps to extract the Wikipedia pages of filtered entities. Sections and subsections follow a standard structure in Wikipedia text. We extract sections and subsections from the text. Text in containers with a depth greater than two is merged into parent sub-sections. 

We also extract the citation URLs in each section using wiki markup. 
We use the \verb|MediaWikiParserFromHell|\footnote{https://pypi.org/project/mwparserfromhell/} module in Python to clean all the wiki markup in a particular section and gather clean section text. We filter the URLs to remove file formats other than HTML and pdf. For each reference URL, we use \verb|BeautifulSoup|\footnote{https://pypi.org/project/beautifulsoup4/} in Python to scrape the <p> paragraph text from the corresponding webpages, and \verb|pdfminer|\footnote{https://pypi.org/project/pdfminer/} to extract the text from pdf. 
Finally, we tokenize the scraped text into individual sentences using a universal sentence tokenizer in IndicNLP~\cite{kakwani2020indicnlpsuite}.
We retain only those sections as part of the dataset which has at least one (crawlable) reference URL with non-empty text. 

Overall, each sample in the dataset consists of the domain, language, section title, set of reference URLs, and Wikipedia section text. This dataset is then split into train, validation, and test in the 60:20:20 ratio, stratified by domain and language. We make this standard splits publicly available as part of the dataset.


\subsection{Data Analysis}
We analyze our prepared dataset across several parameters, the details of which are in the following tables. Table~\ref{tab:article_count} shows the total number of articles per domain per language in the \data{} dataset. By the nature of spread of Wikipedia articles across domains, the number of articles differ across domains per language. Overall, there are $\sim$69K articles from which we extract sections for the dataset.

\setlength{\tabcolsep}{2pt}
\begin{table*}
\centering
\small
\begin{tabular}{|l|r|r|r|r|r|r|r|r||r|}
\hline
\textbf{Domain/Lang}      & \multicolumn{1}{l|}{\textbf{bn}} & \multicolumn{1}{l|}{\textbf{hi}} & \multicolumn{1}{l|}{\textbf{ml}} & \multicolumn{1}{l|}{\textbf{mr}} & \multicolumn{1}{l|}{\textbf{or}} & \multicolumn{1}{l|}{\textbf{pa}} & \multicolumn{1}{l|}{\textbf{ta}} & \multicolumn{1}{l||}{\textbf{en}} & \multicolumn{1}{l|}{\textbf{Total}} \\ \hline
\textbf{Books} & 313 & 922 & 458 & 87 & 73 & 221 & 493 & 1467 & 4034 \\ \hline
\textbf{Film} & 1501 & 1025 & 2919 & 480 & 794 & 421 & 3733 & 1810 & 12683 \\ \hline
\textbf{Politicians} & 2006 & 3927 & 2513 & 988 & 1060 & 1123 & 4932 & 1628 & 18177 \\ \hline
\textbf{Sportsmen} & 5470 & 6334 & 1783 & 2280 & 319 & 1975 & 2552 & 919 & 21632 \\ \hline
\textbf{Writers} & 1603 & 2024 & 2251 & 784 & 498 & 2245 & 1940 & 714 & 12059 \\ \hline
\hline
\textbf{Total}&10893&14232&9924&4619&2744&5985&13650&6538&\textbf{68585}\\
\hline
\end{tabular}
\caption{\data{}: Total \#articles per domain per language}
    \label{tab:article_count}
\end{table*}

\begin{table*}
\centering
\small
\begin{tabular}{|l|r|r|r|r|r|r|r|r|r|}
\hline
\textbf{Domain/Lang}      & \multicolumn{1}{l|}{\textbf{bn}} & \multicolumn{1}{l|}{\textbf{hi}} & \multicolumn{1}{l|}{\textbf{ml}} & \multicolumn{1}{l|}{\textbf{mr}} & \multicolumn{1}{l|}{\textbf{or}} & \multicolumn{1}{l|}{\textbf{pa}} & \multicolumn{1}{l|}{\textbf{ta}} & \multicolumn{1}{l|}{\textbf{en}} & \multicolumn{1}{l|}{\textbf{Total}} \\ \hline
\textbf{Books}       & 434   & 987   & 557   & 111   & 88    & 238   & 598   & 2972 & 5985   \\ \hline
\textbf{Film}        & 2139 & 1363 & 3737 & 676   & 1351 & 476   & 4781 & 4766 & 19289 \\ \hline
\textbf{Politicians} & 3261 & 4478 & 3719 & 1384 & 1404 & 1524 & 6431 & 4780 & 26981 \\ \hline
\textbf{Sportsmen}   & 9485 & 8118 & 2642 & 3056 & 485   & 2624 & 3769 & 2698 & 32877 \\ \hline
\textbf{Writers}     & 2598 & 2743 & 3435 & 1166 & 896   & 3034 & 3113 & 2409 & 19394 \\ \hline
\textbf{Total}& 17917&17689&14090&6393&4224&7896&18692&17625&\textbf{104526}\\
\hline
\end{tabular}
\caption{\data{}: Total \#sections per domain per language}
    \label{tab:section_count}
\end{table*}

Next, Table~\ref{tab:section_count} shows the distribution of number of sections across various (domain, language) pairs in the \data{} dataset.
Further, as mentioned earlier, \data{} is a multi-document summarization dataset. Table~\ref{tab:avg_ref_per_section} shows the average number of references per section for each (domain, language) pair. As can be seen from the table, the dataset contains at least two references on average for every (domain, language) pair, although a large percent of these references are not in the LR language.

\begin{table*}
\centering
\begin{tabular}{|l|r|r|r|r|r|r|r|r|}
\hline
\textbf{Domain/Lang}  & \multicolumn{1}{l|}{\textbf{bn}} & \multicolumn{1}{l|}{\textbf{hi}} & \multicolumn{1}{l|}{\textbf{ml}} & \multicolumn{1}{l|}{\textbf{mr}} & \multicolumn{1}{l|}{\textbf{or}} & \multicolumn{1}{l|}{\textbf{pa}} & \multicolumn{1}{l|}{\textbf{ta}} & \multicolumn{1}{l|}{\textbf{en}} \\ \hline
\textbf{Books}&3.62&2.61&2.59&2.07&3.46&2.30&2.40&6.34\\
\hline
\textbf{Film}&4.85&7.14&3.34&2.96&3.81&4.10&3.83&12.74\\
\hline
\textbf{Politicians}&4.98&4.09&3.75&3.87&2.07&3.59&3.91&14.21\\
\hline
\textbf{Sportsmen}&6.37&8.30&6.96&4.20&3.93&4.49&6.38&21.88\\
\hline
\textbf{Writers}&5.20&5.46&4.16&3.74&2.85&3.34&4.20&17.61\\
\hline
\end{tabular}
\caption{\data{}: Average number of references per section for each domain and language}
\label{tab:avg_ref_per_section}
\end{table*}

\begin{figure}[!b]
    \centering
    \includegraphics[width=\columnwidth]{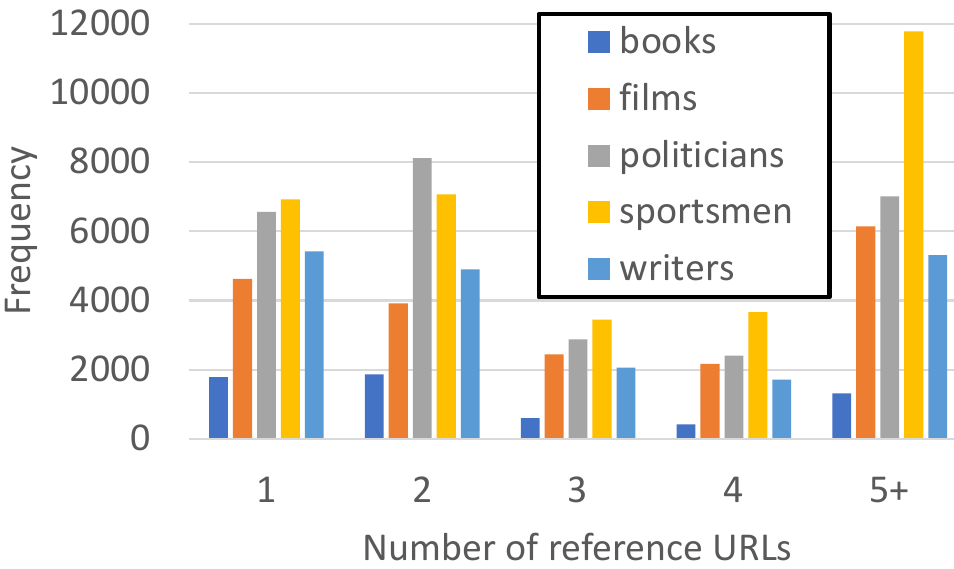}
    \caption{Distribution of number of reference URLs across domains in our \data{} dataset}
    \label{fig:citationDist}
\end{figure}

Fig.~\ref{fig:citationDist} shows the distribution of the number of reference URLs across domains in the dataset. The figure shows that there are several samples where the number of reference URLs is 5+ across all domains showing that multi-document summarization is essential. 

\begin{figure*}
    \centering
    \includegraphics[width=0.88\textwidth]{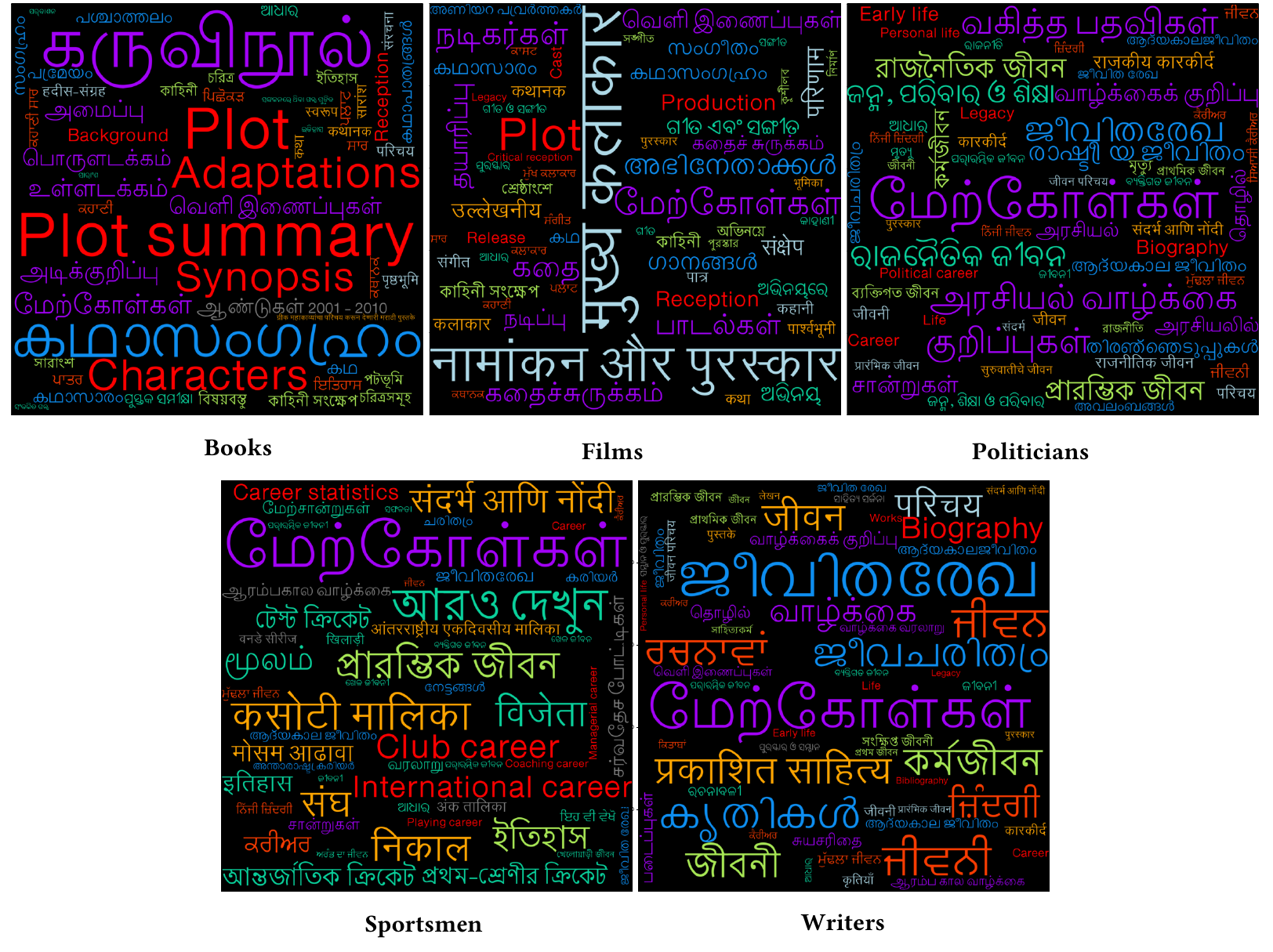}
    \caption{Word clouds of most frequent Wikipedia section titles per domain. Each word cloud contains titles across all languages. Section titles for one language are shown using a single color. Font size indicates relative frequency.}
     \label{fig:wordClouds}
\end{figure*}

Finally, we show word clouds of the most frequent Wikipedia section titles for each of the five domains in Fig.~\ref{fig:wordClouds}. Each word cloud contains the five most frequent titles per language. Section titles for one language are shown using a single color. Font size indicates relative frequency. The word clouds show the variety of section titles per (language, domain) pair.

\section{Two-Stage Approach for \task{}}
\label{sec:approach}
In this section, we first motivate the need for proposing a two-stage approach for the cross-lingual multi-document summarization task, \task{}. Next, we discuss the details of the two stages: extractive and abstractive. Finally, we present multiple training setups.

Table~\ref{tab:avg_sents_refs} shows the average number of sentences in references of a section for each domain and language in our dataset. Combined with the number of references per section as shown in Table~\ref{tab:avg_ref_per_section}, the overall text input is very large. Given the quadratic complexity of Transformer-based methods, it is infeasible to feed such long inputs to an encoder-decoder model and expect it to be able to output reasonable summaries. Transformers sub-quadratic complexity is an active area of research with models like Longformer~\cite{beltagy2020longformer}, Reformer~\cite{kitaev2019reformer}, etc. But we plan to explore them as part of future work.

\begin{table*}[hbt!]
\small
\centering
\begin{tabular}{|l|r|r|r|r|r|r|r|r|}
\hline
\textbf{Domain}      & \multicolumn{1}{l|}{\textbf{bn}} & \multicolumn{1}{l|}{\textbf{hi}} & \multicolumn{1}{l|}{\textbf{ml}} & \multicolumn{1}{l|}{\textbf{mr}} & \multicolumn{1}{l|}{\textbf{or}} & \multicolumn{1}{l|}{\textbf{pa}} & \multicolumn{1}{l|}{\textbf{ta}} & \multicolumn{1}{l|}{\textbf{en}} \\ \hline
\textbf{Books}&200.2&117.9&1232.0&225.8&51.9&246.7&302.7&940.8\\
\hline
\textbf{Films}&223.9&320.6&91.9&105.6&345.9&172.6&192.5&1253.6\\
\hline
\textbf{Politicians}&1318.3&467.1&513.3&394.0&54.5&255.4&614.1&1540.9\\
\hline
\textbf{Sportsmen}&335.7&1166.3&406.9&167.5&724.0&253.5&714.0&1535.0\\
\hline
\textbf{Writers}&643.2&2032.5&800.1&385.5&118.5&351.0&1279.0&2061.3\\
\hline
\end{tabular}
\caption{Average number of sentences in references of a section for each domain and language in \data{}.}
\label{tab:avg_sents_refs}
\end{table*}

In order to address the long input problem, we propose a two-stage system where the first stage identifies promising candidate sentences across all the reference citations for a sample. The highest scoring candidate sentences are passed as input to the second stage which generates an abstractive summary. In the following, we will discuss the two stages in detail.



\subsection{Extractive Summarization Stage}

Given a set of reference URLs, the extractive stage aims at selecting a subset of sentences from these URLs which best represents a summary of the set of URLs. While earlier methods for extractive summarization were position based, or lexical chains based, neural methods have become popular in the past decade. We experiment with two different extractive summarization-based techniques: Salience and HipoRank. For both methods, the input consists of the section title and a set of reference URLs. Both the methods output a summary worthiness score for every sentence in these reference URLs.




\noindent\textbf{Salience based extractive summarization}: 
The main idea of salience based extractive summarization is to find the top-K salient sentences from the input references based on the relevance of that sentence relative to a particular section title. Our salience method is inspired by the relevance scoring method in~\cite{yasunaga2021qa}, where a language model was used to calculate the relevance score of each answer entity relative to the QA (question-answer) context. For extracting the top-K sentences, we first split the reference text into sentences. Each sentence is then prepended with a section title and passed as input to a pretrained XLM-RoBERTa\cite{conneau2019unsupervised} language model. We score each sentence based on the likelihood from the language model. Top-K sentences with the highest relevance scores are passed on as output to the next stage.

Note that we use a pretrained language model for this method. The model is used in probe mode only.


\noindent\textbf{HipoRank based extractive summarization}: 
Hierarchical and Positional Ranking model (HipoRank)~\cite{dong2020discourse} is an unsupervised graph-based model for the extractive summarization of long documents. Given a document with multiple sections, it creates a directed hierarchical graph with sentence and section nodes and sentence-sentence and sentence-section edges with asymmetrically weighted edges. Score for a sentence node is then computed based on a weighted sum of edges incident on the node.

We compute sentence node representations using mBERT~\cite{devlin2018bert}. We take the mean of all the sentence representations within a section to compute the representation for every section node.

Each sentence node is connected to other nodes via intra-sectional and inter-sectional edges. Intra-sectional connections are between all sentences of the same section, meant to model the local importance of the sentence. The key idea is that sentences similar to most sentences within a section are more important. On the other hand, inter-sectional connections are between sentences and section nodes, meant to model the global importance of the sentences. Here, the idea is that sentences most similar to other sections are the most important. For efficiency, edges are not allowed between two sentences in different sections. 

Cosine similarity between node embeddings is used to compute edge weights. Based on the hypothesis that important sentences are near the boundaries (start or end) of a text, intra-sectional edges have higher weight if they are incident on a boundary sentence. Similarly, important sections are near the boundaries of the document. This hypothesis is used to weigh inter-sectional edges appropriately. Finally, the importance score for a sentence node is computed based on a weighted sum of edges (both intra-sectional as well as inter-sectional) incident on the node. We then sort these sentences in descending order based on the importance score and greedily select the top-K sentences as our extractive summary.


\subsection{Abstractive Summarization Stage}
Note that the output from the extractive stage is in the reference text language itself. Also, since these sentences have been obtained across several documents, they often form an incoherent extractive summary. We need an abstractive stage to generate coherent summaries in the target language. For the abstractive stage, we use two state-of-the-art multi-lingual natural language generation models viz. mBART-large\cite{liu2020multilingual} and mT5-base\cite{xue2021mt5}. mT5 and mBART are both multi-lingual encoder-decoder Transformer models and have been shown to be very effective across multiple such NLP tasks like question answering, natural language inference, named entity  recognition, etc. Both these models contain 24 layers (12 layers encoder + 12 layers decoder). For both models, we pass the target language id, article title, section title, and top-k sentences from the extractive stage (descending sorted based on score) as input. 


mT5~\cite{xue2021mt5} was pretrained on mC4 dataset\footnote{\url{https://www.tensorflow.org/datasets/catalog/c4\#c4multi-lingual_nights_stay}} comprising of web data in 101 different languages and leverages a unified text-to-text format. mBART~\cite{liu2020multilingual} was pretrained on CommonCrawl corpus using the BART objective where the input texts are noised by masking phrases and permuting sentences, and a single Transformer model is learned to recover the texts. Specifically, our mT5-base model is an encoder-decoder model with 12 layers each for the encoder and decoder. It has 12 heads per layer, a feed-forward size of 2048, keys and values are 64 dimensional, $d_{model}$=768, and a vocabulary size of 250112. Overall the model has 582.40M parameters. Our mBART-large-50 model~\cite{liu2020multilingual} also has 12 layers each for encoder as well as decoder. It has 16 heads per layer, a feed-forward size of 4096, $d_{model}$=1024, and a vocabulary size of 250054. Overall the model has 610.87M parameters. Note that the two models have almost the same size.

    

Using the training part of our \data{} dataset, we fine-tune both these models on the extractive stage output.

\subsection{Multi-lingual, Multi-domain, and Multi-lingual-Multi-domain setups}
\data{} contains data for eight languages and five domains. We could perform training in various ways. We could train one model per (language, domain) pair. Given five domains and eight languages, we would need to train, maintain and deploy 40 models. Also, the amount of training data per (language, domain) pair is not very large. Thus, such individual models may not be able to benefit from cross-language or cross-domain knowledge.

Another way of training models is multi-lingual. This means that we train one model per domain using training data across all languages. Thus, there will be five models. A third way is to training models in a multi-domain manner. Thus, we will have one model per language using training data across all domains, leading to eight models.

One last approach is to train a multi-lingual-multi-domain model. We collate training data across all languages and domains and train a single model. This model can exploit cross-language cross-domain clues and learn robust representations.

Previous literature in multi-lingual cross-lingual natural language generation has shown that multi-lingual models tend to be better than individual ones, especially for low-resource languages. Since this work is focused on LR languages,  we experiment with multi-lingual, multi-domain, and multi-lingual-multi-domain setups.

\section{Experiments}
\label{sec:expts}
\subsection{Training Configuration}

The two stages in our approach have different compute  requirements. We performed extractive step on a machine with one NVIDIA 2080Ti with 12GB of GPU RAM. For the abstractive stage, we fine-tuned the model on a machine having NVIDIA V100 having 32GB of GPU RAM with CUDA 11.0 and PyTorch 1.7.1.


For the salience based extractive stage, we used XLM-RoBERTa-base\cite{conneau2019unsupervised} model for extracting the sentence representation with 512 as the maximum input length. For HipoRank, we used the multi-lingual BERT (mBERT~\cite{devlin2018bert}) model to get the sentence representation for building the graph with 512 as maximum input length. We took a maximum of 50 sentences per sample as output from the extractive stage.

\setlength{\tabcolsep}{2pt}
\begin{table*}[!hbt]
    \centering
    \small
    \begin{tabular}{|l|l|l|c|c|c|}
    \hline
&Extractive&Abstractive&ROUGE-L&chrF++&METEOR\\
    \hline
\hline
\multirow{4}{*}{\parbox{1cm}{Multi-lingual}}&Salience&mBART&15.59&17.20&10.98\\
\cline{2-6}
&Salience&mT5&14.66&15.45&8.92\\
\cline{2-6}
&HipoRank&mBART&\textbf{16.96}&\textbf{19.11}&\textbf{12.19}\\
\cline{2-6}
&HipoRank&mT5&15.98&17.11&10.08\\
\hline
\multirow{4}{*}{\parbox{1cm}{Multi-domain}}&Salience&mBART&\textbf{19.88}&\textbf{22.82}&\textbf{15.00}\\
\cline{2-6}
&Salience&mT5&12.13&13.66&7.27\\
\cline{2-6}
&HipoRank&mBART&18.87&20.79&14.10\\
\cline{2-6}
&HipoRank&mT5&12.29&13.93&7.36\\
\hline
\multirow{4}{*}{\parbox{1cm}{Multi-lingual-multi-domain}}&Salience&mBART&20.50&22.32&14.81\\
\cline{2-6}
&Salience&mT5&17.31&18.77&11.57\\
\cline{2-6}
&HipoRank&mBART&\underline{\textbf{21.04}}&\underline{\textbf{23.44}}&\underline{\textbf{15.35}}\\
\cline{2-6}
&HipoRank&mT5&17.65&19.04&11.74\\
\hline
    \end{tabular}
    \caption{\task{} Results across multiple training setups and (extractive, abstractive) methods on test part of \data{}. Best results per block are highlighted in bold. Overall best results are also underlined.}
    \label{tab:mainResults}
\end{table*}

For the abstractive stage, we fine-tuned mBART\cite{liu2020multilingual} and mT5\cite{xue2021mt5} models for 20 epochs keeping a batch size of 4. We initialize using google/mt5-base and facebook/mbart-large-50 huggingface checkpoints. We kept the maximum input and output length as 512 across all of our experiments. We used AdamW optimizer with a learning rate of 1e-5. We perform greedy decoding.

\subsection{Metrics}
We evaluate our models using standard Natural Language Generation (NLG) metrics like ROUGE-L~\cite{lin-2004-rouge}, METEOR~\cite{banerjee-lavie-2005-meteor} and chrF++~\cite{popovic2017chrf++}. Another popular NLG metric is PARENT. But PARENT~\cite{dhingra2019handling} relies on the word overlap between input and the prediction text. Since the input and prediction in \task{} are in different languages, we cannot compute PARENT scores. 

\begin{enumerate}
    \item \textbf{ROUGE-L}: ROUGE-L, or Recall-Oriented Understudy LCS is based on statistics using the longest common subsequence (LCS). The longest common subsequence task automatically determines the longest co-occurring n-grams given a reference sequence and a machine-generated sequence, while taking sentence-level structure similarities into account.
    \item \textbf{chrf++}: In addition to adding word n-grams, chrF++ is an evaluation measure that uses the F-score statistic for character n-gram matches.
    \item \textbf{METEOR}: METEOR, an automated metric evaluation, is based on a generalized idea of unigram matching between the text generated by the machine and the reference text created by a human. Based on their meanings, surface forms, and stemmed forms, unigrams can be matched.
\end{enumerate}

\section{Results}
\label{sec:results}

Table~\ref{tab:mainResults} shows results across two extractive methods (salience, HipoRank), two abstractive methods (mBART, mT5), three training setups (multi-lingual, multi-domain, multi-lingual-multi-domain), and three metrics (ROUGE-L, METEOR, and chrF++) computed as a micro-average across all test instances in \data{}.

\begin{table*}[!t]
    \centering
    \small
    \begin{tabular}{|l|c|c|c|c|c|c|c|c|}
    \hline
&bn&en&hi&mr&ml&or&pa&ta\\
\hline
\hline
ROUGE-L & 14.49 & 7.46 & 29.01 & 20.67 & 12.25 & 25.54 & 16.89 & 17.09 \\ \hline
chrF++ & 18.58 & 10.55 & 28.38 & 20.41 & 15.30 & 27.31 & 13.49 & 21.90 \\ \hline
METEOR & 9.71 & 5.90 & 25.24 & 13.72 & 6.42 & 22.69 & 10.12 & 9.87 \\ \hline
    \end{tabular}\\
    \textbf{Multi-lingual HipoRank+mBART}\\
       \begin{tabular}{|l|c|c|c|c|c|c|c|c|}
    \hline
&bn&en&hi&mr&ml&or&pa&ta\\
\hline
\hline
ROUGE-L&15.30&12.07&36.16&31.25&14.22&29.53&16.91&15.00\\
\hline
chrF++&19.40&17.41&34.34&32.50&18.34&32.20&14.10&21.65\\
\hline
METEOR&10.34&9.59&31.02&24.86&8.89&26.86&10.01&9.29\\
\hline
        \end{tabular}\\
    \textbf{Multi-domain Salience+mBART}\\
          \begin{tabular}{|l|c|c|c|c|c|c|c|c|}
            \hline
&bn&en&hi&mr&ml&or&pa&ta\\
\hline
\hline
ROUGE-L & 15.21 & 16.32 & 36.38 & 22.71 & 15.50 & 27.41 & 18.64 & 18.87 \\ \hline
chrF++ & 19.50 & 21.34 & 34.55 & 21.93 & 18.65 & 28.83 & 16.27 & 23.99 \\ \hline
METEOR & 10.24 & 12.74 & 31.24 & 14.88 & 8.84 & 23.93 & 11.6 & 11.26 \\ \hline
        \end{tabular}\\
    \textbf{Multi-lingual-multi-domain HipoRank+mBART}
    \caption{Detailed per-language results on test part of \data{}, for the best model per training setup.}
    \label{tab:perLangResults}
\end{table*}

\begin{table*}[hbt!]
    \centering
    \small
    \begin{tabular}{|l|c|c|c|c|c|}
    \hline
&writers&books&sportsmen&politicians&films\\
\hline
\hline
ROUGE-L&10.12&3.65&20.61&22.01&14.60\\
\hline
chrF++&10.76&3.58&22.94&24.34&18.36\\
\hline
METEOR&5.77&1.93&14.66&17.61&10.04\\
\hline
    \end{tabular}\\
    \textbf{Multi-lingual HipoRank+mBART}\\
        \begin{tabular}{|l|c|c|c|c|c|}
    \hline
    &writers&books&sportsmen&politicians&films\\
\hline
\hline
ROUGE-L & 14.21 & 20.17 & 20.65 & 22.77 & 20.82 \\ \hline
chrF++ & 17.24 & 21.86 & 22.75 & 26.14 & 24.30 \\ \hline
METEOR & 10.06 & 16.26 & 14.71 & 18.88 & 14.81 \\
\hline
        \end{tabular}\\
    \textbf{Multi-domain Salience+mBART}\\
            \begin{tabular}{|l|c|c|c|c|c|}
            \hline
                &writers&books&sportsmen&politicians&films\\
\hline
\hline
ROUGE-L&14.67&22.03&20.44&23.70&21.60\\
\hline
chrF++&16.65&22.81&21.57&25.75&24.51\\
\hline
METEOR&9.81&17.55&13.84&18.92&15.11\\
\hline
        \end{tabular}\\
    \textbf{Multi-lingual-multi-domain HipoRank+mBART}\\
    \caption{Detailed per-domain results on test part of \data{}, for the best model per training setup.}
    \label{tab:perDomainResults}
\end{table*}

\setlength{\tabcolsep}{1.5pt}
\begin{table*}
    \centering
    \small
    \begin{tabular}{|l|c|c|c|c|c|c|c|c|c|c|c|c|c|c|c|}
    \hline
&\multicolumn{5}{c|}{ROUGE-L}&\multicolumn{5}{c|}{chrf++}&\multicolumn{5}{c|}{METEOR}\\
\hline
&writers&books&sportsmen&politicians&films&writers&books&sportsmen&politicians&films&writers&books&sportsmen&politicians&films\\
\hline
\hline
bn&10.61&9.43&15.78&17.46&15.75&14.72&14.19&20.28&21.21&20.03&6.13&5.66&10.56&12.99&10.39\\
\hline
en&13.04&15.62&18.53&13.32&20.15&19.71&18.90&22.80&20.00&24.13&10.65&11.62&13.89&11.47&15.09\\
\hline
hi&33.23&58.71&28.48&53.18&21.46&31.05&51.99&26.99&52.05&19.64&28.49&53.78&21.46&51.65&15.30\\
\hline
mr&15.37&17.00&26.77&20.06&24.15&14.68&16.24&26.84&18.12&21.82&7.40&9.50&20.14&10.74&14.30\\
\hline
ml&8.96&10.93&12.97&14.36&24.19&13.35&12.18&15.42&18.01&26.51&3.92&4.77&6.14&7.73&16.16\\
\hline
or&13.15&12.31&9.38&43.76&26.66&14.44&15.16&10.51&44.17&29.27&5.67&9.14&5.28&40.89&23.30\\
\hline
pa&14.96&12.35&24.54&16.59&17.15&13.42&12.39&21.32&14.02&13.82&8.59&7.48&16.54&9.80&9.63\\
\hline
ta&10.62&11.85&18.94&19.18&24.90&16.43&17.63&23.98&23.77&29.94&4.89&6.29&10.03&11.24&17.05\\
\hline
    \end{tabular}
    \caption{Detailed results for every (domain, language) partition of the test set of our \data{} dataset, for our best \task{} model: Multi-lingual-multi-domain HipoRank+mBART.}
    \label{tab:fullDetailedResults}
\end{table*}

\begin{figure*}
    \centering
    \includegraphics[width=0.9\textwidth]{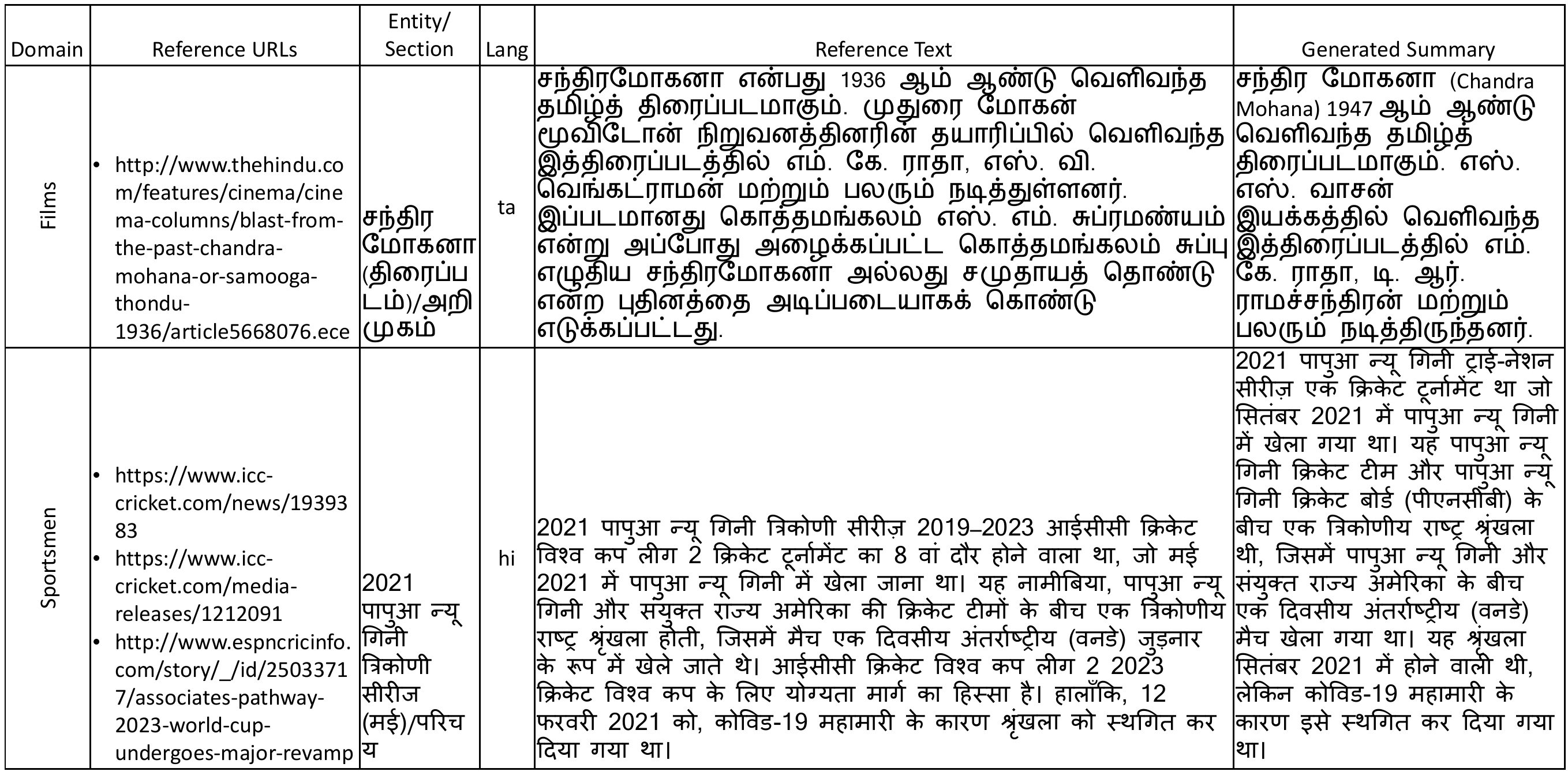}
    \captionof{table}{Some examples of \task{} using our best model. More examples are in Appendix.}
    \label{tab:caseStudies}
\end{figure*}

From the table, we observe that best results are obtained using the multi-lingual-multi-domain training setup. Also, in this setup, the combination of HipoRank with mBART provides the best overall results. These results are statistically significantly better compared to other rows in the table. The supremacy of the multi-lingual-multi-domain training setup is expected given that it combines learning across all languages and domains in the dataset. Also, HipoRank was expected to perform better since it combines the knowledge of the pretrained (mBERT) model with the hierarchical document structure. Even for the multi-lingual setup, best results are obtained using the HipoRank+mBART combination. However, for the multi-domain setup, we observe that Salience+mBART performs better. Mann-Whitney U-Test shows that, amongst the best block (Multi-lingual-multi-domain), our best model (Hiporank+mBART) is better than the second best (Salience+mBART) with p-values of 5.47e-07 (ROUGE-L), 2.76e-26 (chrF++), and 6.76e-09 (METEOR). Also Hiporank+mBART is better than the worst model in the  block (Salience+mT5) with p-values of 3.16e-143 (ROUGE-L), 1.69e-268 (chrF++) and 8.67e-201 (METEOR). Also, note that Salience uses XLM-RoBERTa (270M parameters); HipoRank uses mBERT (110M parameters). mT5 and mBART contain 580M and 610M parameters resp. Finally, Average output length is 221 words for our best model. 


Further, we wish to drill deeper into the performance of the best models for each of the training setups. Hence, for these three models, we show micro-averaged metrics per language and per domain for the test set in Tables~\ref{tab:perLangResults} and~\ref{tab:perDomainResults} respectively. We make the following observations from Table~\ref{tab:perLangResults}: (1) Multi-domain training is much better than multi-lingual training except for Tamil (ta). (2) Interesting relatively richer languages like en and hi seem to benefit most when we move from multi-lingual to multi-lingual-multi-domain setup. (3) When comparing multi-domain training with multi-lingual-multi-domain, we observe gains across most languages except for losses in mr and or. From Table~\ref{tab:perDomainResults}, we observe that across all domains, results improve as we move from multi-lingual training to multi-domain training to multi-lingual-multi-domain setup (except for minor drop for sportsmen in the multi-lingual-multi-domain case).



Finally, we present the most detailed per (domain, language) level results for our best model in Table~\ref{tab:fullDetailedResults}. We observe that best results are obtained for the hi-books combination. Overall, the model works best for Hindi across all domains. The model also performs reasonably for mr and or. But more work needs to be done to improve the model for Bengali and Malayalam.



For a qualitative analysis of our best model outputs, we show some sample outputs in Table~\ref{tab:caseStudies}. In general, our model generates fluent text to a certain length. But, as the length of the output grows, we see the repeated patterns in the text, breaking the sentence structure. Pretrained language models usually present this problem of repeating n-grams, and increasing the training dataset size has been shown to alleviate it.
Further, we observe the faithfulness of content between the generated text and reference text. Despite generating correct sentence structure, the model is seen to predict value strings incorrectly, like that of date of birth, names of persons, and related entities. This issue of hallucination is also common in pretrained language models and finetuning on more training data should help.

\section{Conclusion}
\label{sec:conclusions}
We motivated the need for cross-lingual multi-document summarization for generating Wikipedia text for low-resource languages. Toward this \task{} task, we contribute a novel dataset spanning five domains, eight languages and $\sim$105K summaries. We also proposed a two-stage extractive-abstractive system and experimented with multiple training setups. We observed that our multi-lingual-multi-domain model using HipoRank (extractive) and mBART (abstractive) led to the best results. We make our code and dataset publicly available and hope that the community will use it as a benchmark to work further in this critical area.

\bibliography{main}
\bibliographystyle{acl_natbib}

\newpage
\appendix
\section{More Examples for Qualitative Analysis}

\begin{figure*}[!t]
    \centering
    \includegraphics[width=\textwidth]{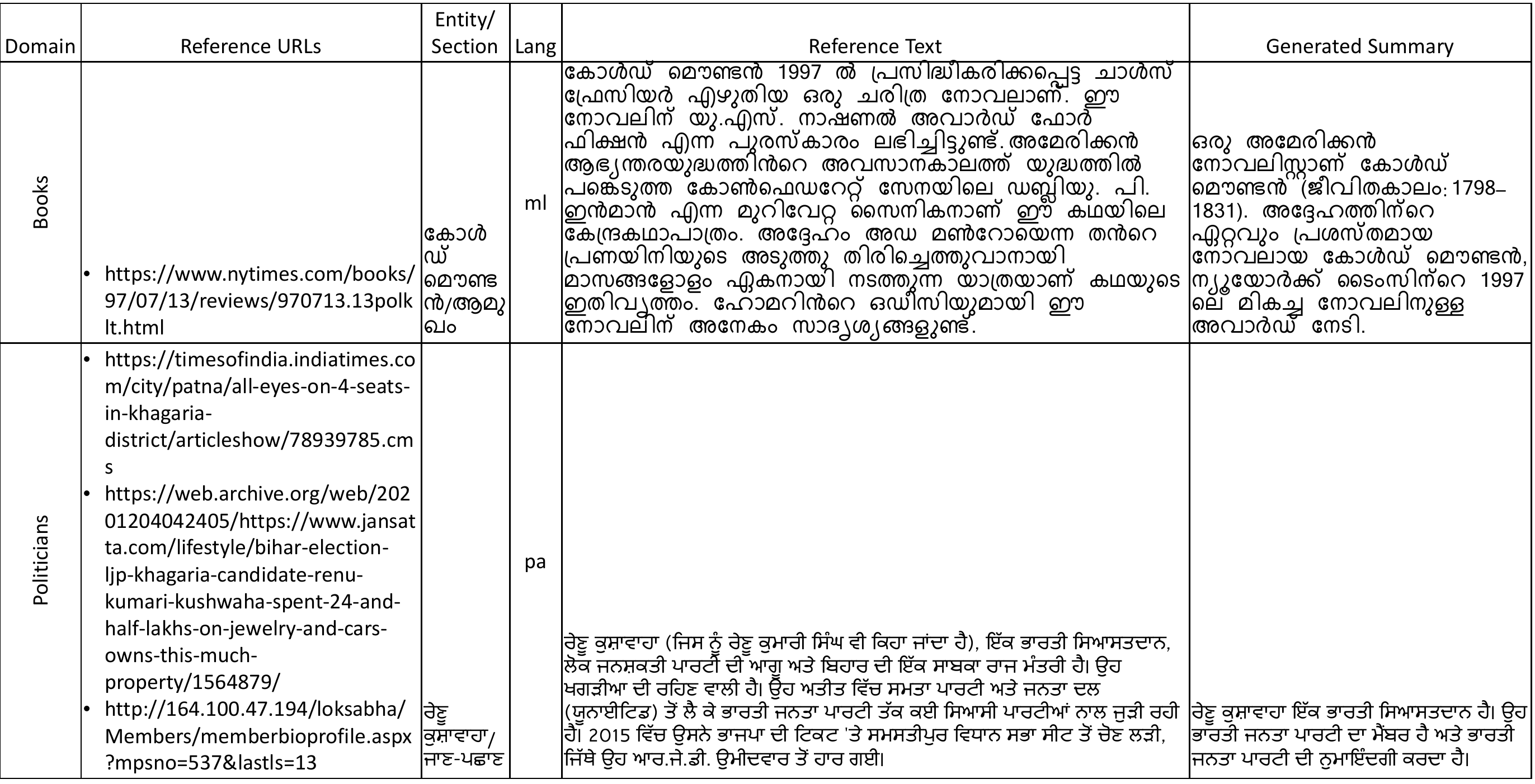}
    \captionof{table}{More examples of \task{} using our best model.}
    \label{tab:caseStudies2}
\end{figure*}

\begin{figure*}[!t]
    \centering
    \includegraphics[width=\textwidth]{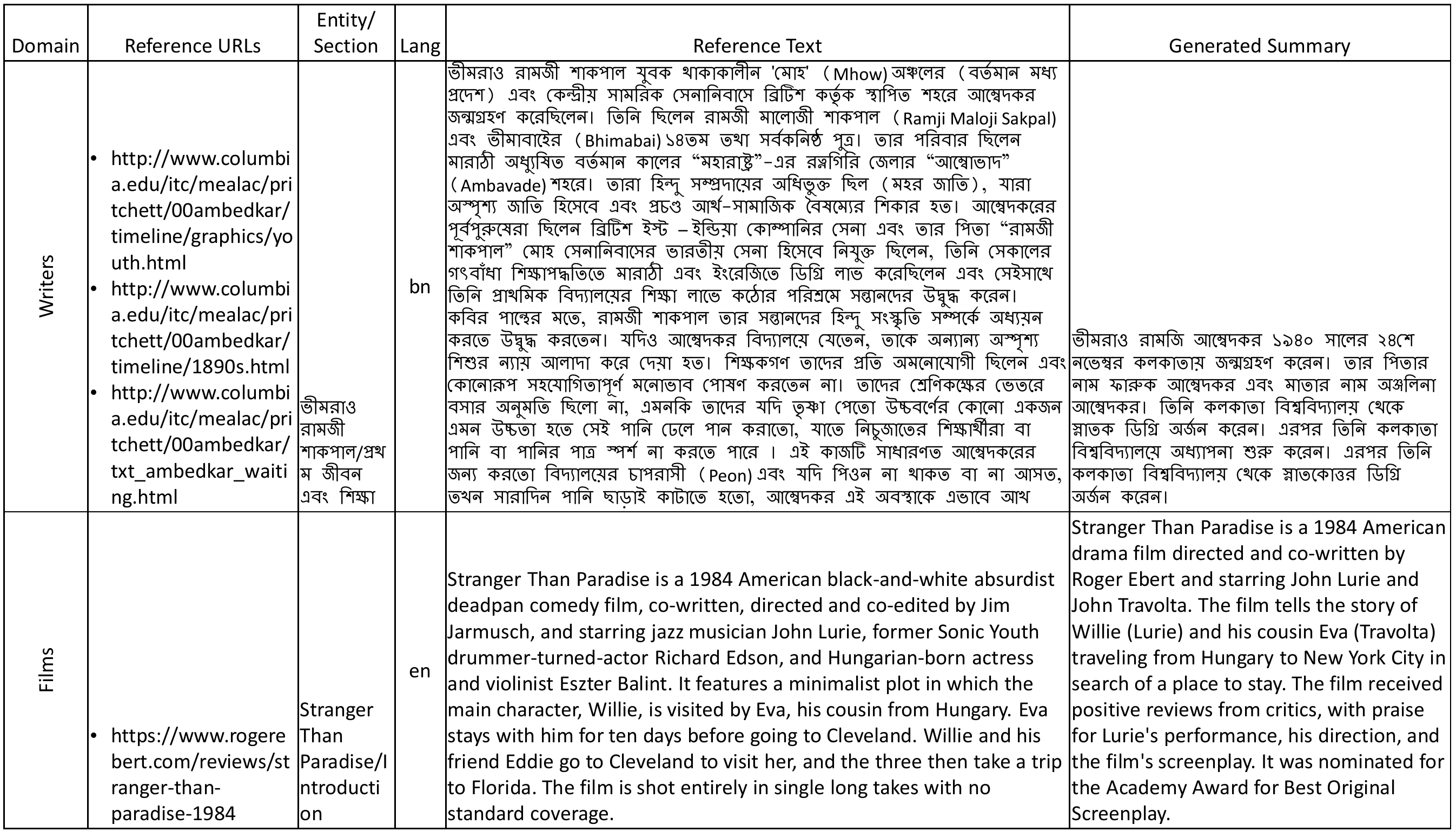}
    \captionof{table}{More examples of \task{} using our best model.}
    \label{tab:caseStudies3}
\end{figure*}

For a qualitative analysis of our best model outputs, we show more sample outputs in Tables~\ref{tab:caseStudies2} and~\ref{tab:caseStudies3}. In general, our model generates fluent text to a certain length. But, as the length of the output grows, we see the repeated patterns in the text, breaking the sentence structure. Pretrained language models usually present this problem of repeating n-grams, and increasing the training dataset size has been shown to alleviate it.
Further, we observe the faithfulness of content between the generated text and reference text. Despite generating correct sentence structure, the model is seen to predict value strings incorrectly, like that of date of birth, names of persons, and related entities. This issue of hallucination is also common in pretrained language models and finetuning on more training data should help.


\end{document}